# DETERMINATION OF PEDESTRIAN FLOW PERFORMANCE BASED ON VIDEO TRACKING AND MICROSCOPIC SIMULATIONS [*]

By Kardi TEKNOMO[**], Yasushi TAKEYAMA[***], Hajime INAMURA[****]

## 1. Introduction

One of the objectives of understanding pedestrian behavior is to predict the effect of proposed changes in the design or evaluation of pedestrian facilities. We want to know the impact to the user of the facilities (i.e. pedestrian), as the design of the facilities change. That impact was traditionally evaluated by level of service standards (LOS). Another design criterion to measure the impact of design change is measured by the pedestrian flow performance index (PI). This paper describes the determination of pedestrian flow performance based video tracking or any microscopic pedestrian simulation models.

Most of pedestrian researches (for best classical examples see [2,8]) have been done on a macroscopic level, which is an aggregation of all pedestrian movement in pedestrian areas into flow, average speed and area module. Macroscopic level, however, does not consider the interaction between pedestrians. It is also not well suited for prediction of pedestrian flow performance in pedestrian areas or in buildings with some obstruction that reduces the effective width of the walkways. On the other hand, the microscopic level has a more general usage and considers detail in the design.

Case studies by microscopic simulation have been reported [3,4] that the flow performance of pedestrian in the intersection of pedestrian malls and doors could be improved by putting some control such as roundabout or direction rule. More efficient pedestrian flow can even be reached with less space. Those results have rejected the linearity assumption of space and flow in the macroscopic level.

Since there are many types of Microscopic Pedestrian Simulation Models (MPSM)[1,3-6], there is a need for a unifying language so that all MPSM techniques can be used interchangeably without any confusion. The NTXY database, as explained in the next section, could take the role of language unification. Based on the NTXY database, flow performance formulation can be determined.

## 2. NTXY Database

Recently, the algorithm to automate the tracking of pedestrian from video files has been developed by the authors [7]. Images of pedestrians are taken by video camera in the field, on top of pedestrian facilities. The images are then converted into files in the laboratory. Normally, one second on a video film consist of many frame pictures. Depending on the number of frame per second, the time recorded for each frame can be determined. Figure 1 shows the example of six consecutive frames using 2 frames per second. All the frame-files that have been gathered can be put together into a single stack-file.

Each image of pedestrian in each frame is represented by one point, which is the centroid of its area. Starting from the first frame, each point is numbered by a unique pedestrian-number. Simple tracking algorithm was performed to follow the movement-path coordinates of each pedestrian for every frame. The objective of the tracking algorithm is to match the points between frames by giving a pedestrian number to each point in each frame. Two points in two consecutive frames are matched if and only if the two points represent one pedestrian. Each pedestrian is traced frame by frame from the time he/she shows up in the pedestrian trap until he/she goes out of the trap. Pedestrian movement database, called NTXY database, is produced from these data collection procedures.

Table 1 shows the example of a NTXY database. The database consists of four fields, which are pedestrian number, time and coordinate location. Pedestrian number is a unique number for each pedestrian and new pedestrian number is given to a new person who enters the pedestrian trap. This number is useful to distinguish the data of a pedestrian from the another. Only pedestrians in the pedestrian trap are recorded. The coordinate location of each pedestrian is the real world coordinate of the pedestrian's image. Each row in the database represents a single observation point. One observation point is the position of a single pedestrian in one frame. The reference point of the coordinate system is arbitrary. For convention, in a common one or two ways traffic flow, the pedestrians are moving in the Y direction.

Time recorded, $T$, represents the frame number. Because every frame is tapped with a constant time interval, $T$ can be called clock time and treated as discreet.

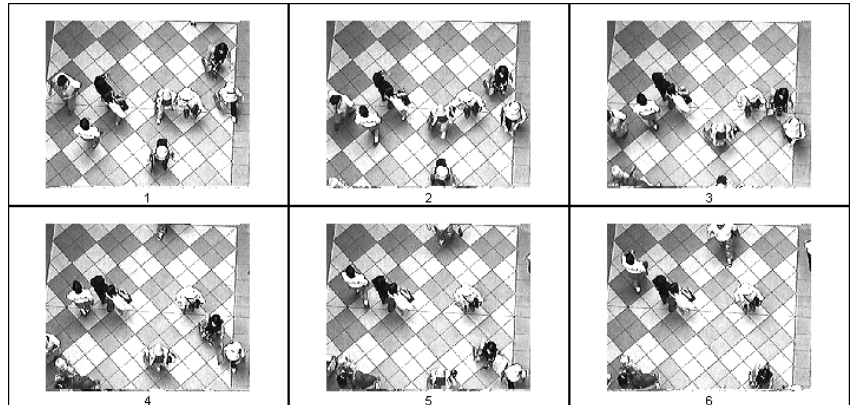

**Figure 1. Example of Six Consecutive Frames**

Similar NTXY database can be created by any Microscopic Pedestrian Simulation Models. Both continuous and discreet based simulator can record the movement data of each pedestrian at every constant interval and write them into a text file.

The NTXY database may function as a design criterion and a unifying language. As can be seen in Figure 2, pedestrian flow performance can be calculated based on NTXY database. Whether the data comes from the real world or from simulation models, the flow performance can be calculated with the same definition of flow performance. If the design of pedestrian facilities is changed (in either real world or simulation), then a new NTXY database can be produced. By comparing the flow performance before and after the design change, a better design can be revealed. In this case, the flow performance functions as a design criterion for the facilities. It gives feedback for better design.

**Table 1. Part of NTXY Database**

| Pedestrian number N | Time T | X | Y |
|---|---|---|---|
| 3 | 13.5 | 475 | 411 |
| 3 | 14.0 | 475 | 429 |
| 3 | 14.5 | 470 | 448 |
| 3 | 15.0 | 469 | 473 |
| 4 | 2.0 | 586 | 145 |
| 4 | 2.5 | 592 | 233 |
| 4 | 3.0 | 598 | 311 |
| 4 | 3.5 | 598 | 379 |
| 4 | 4.0 | 585 | 442 |
| 5 | 2.5 | 258 | 236 |
| 5 | 3.0 | 264 | 266 |
| 5 | 3.5 | 253 | 290 |
| 5 | 4.0 | 250 | 315 |
| 5 | 4.5 | 253 | 337 |
| 5 | 5.0 | 242 | 362 |
| 5 | 5.5 | 230 | 382 |
| 5 | 6.0 | 226 | 408 |
| 5 | 6.5 | 217 | 426 |
| 5 | 7.0 | 201 | 439 |
| 5 | 7.5 | 192 | 458 |
| 5 | 8.0 | 183 | 475 |
| 6 | 1.0 | 311 | 269 |
| 6 | 1.5 | 309 | 299 |

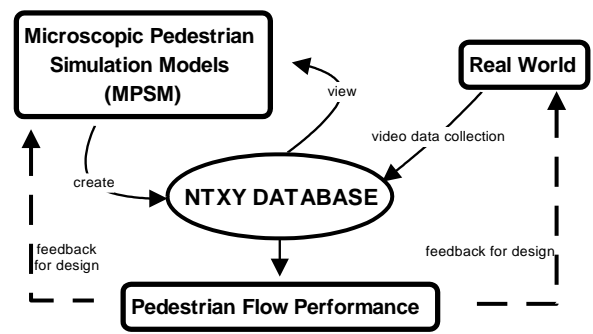

**Figure 2. Function of NTXY Database**

The NTXY database can be used as a standard file of results from any MPSM and microscopic data collection of pedestrians by video. All MPSM can exchange data with another through the NTXY database. One simulator can use the pedestrian movement data produced by another type of simulator. It is a unifying language for all MPSM. Once the NTXY database is created, the movement of pedestrians can be viewed as if in the simulation using the viewer-program. Data collected from the real world also can be viewed as a simulation using the simple viewer-program.

### 3. Flow Performance Determination

In this section flow performance formulation based on NTXY database is determined. The flow performance consisting of delay and uncomfortability can be aggregated into a single value of Performance Index (PI). From the movement of an individual pedestrian at time $t$, individual uncomfortability index and delay is obtained and later aggregated for a certain time interval to acquire the Performance Index. Let us assume that pedestrian number $\alpha$ starts with number 1 and increases in order ($\alpha = 1, 2, 3, \ldots$). In the NTXY database, the observation time is constant. The coordinate location of pedestrian $\alpha$ at time $t$ is $\bar{x}_{\alpha_t}$ and $\rho_\alpha$ is the total number of observations of pedestrian $\alpha$ and $ti_\alpha$ and $to_\alpha$ are the time when the pedestrian $\alpha$ is first recorded and last recorded, respectively. Assuming that the most comfortable walking displacement is uniform and in a straight line, the variance of walking displacement, $\bar{\gamma}_\alpha$, is the measurement of the square deviation of walking displacement toward it ideal displacement. In other words, it measures uncomfortable walking displacement, so it called individual uncomfortability index. Since the time interval between walking displacement is constant, the uncomfortability index also represents the speed and direction change. Individual delay is the difference between the real travel time of pedestrian in the pedestrian trap and the travel time of straight-line distance by the average speed, divided by the walking distance. The unit can be second per meter. Table 2 shows the formulation of individual flow performance.

Performance of the pedestrian facilities can be determined by measuring pedestrian delay and uncomfortability. Individual performance index is a set of a linear combination of delay and uncomfortability index.

$$PI_\alpha = a.\gamma_\alpha + b.\lambda_\alpha \tag{1}$$

Where $a$ and $b$ are weight factor between uncomfortability and delay.

Aggregation of the individual flow performance for all individuals is performed within a certain time interval, from $T_1$ until $T_2$. If the total number of pedestrians observed during time interval $T_1$ until $T_2$ is

$$n = \sum_{t=T_1}^{T_2} \alpha_t \tag{2}$$

Then the Pedestrian Performance Index is the average of all individual Performance Index.

$$PI = \frac{\sum_\alpha PI_\alpha}{n} \tag{3}$$

The lower the value of the Pedestrian Performance Index, the better design the pedestrian facility has.

### 4. Traffic Flow Variables

Apart from flow performance, the macroscopic flow variables such as flow, time mean speed, space mean speed, area module and direction can also be calculated based on the NTXY database as formulated in table 3.

### 5. Conclusion

A new insight for performance measurement as an aggregation of pedestrian interaction is proposed. Pedestrian Performance Index that measures the delay and uncomfortability can be used for evaluation of pedestrian facilities using before and after studies, or microscopic pedestrian simulation model. The determination of flow performance is based on the NTXY database, which is the bridge between microscopic pedestrian simulation model, video data collection and pedestrian flow performance. Both simulator and the result of image processing *create* the NTXY database and it is useful as their unifying language.

**Table 2. Individual Pedestrian Flow Performance**

| Flow Performance | Formula |
|---|---|
| Walking distance at time t | $d_{\alpha_t} = \|\vec{X}_{\alpha_{t+\Theta}} - \vec{X}_{\alpha_t}\|$ |
| Walking displacement at time t | $\vec{d}_{\alpha_t} = \vec{X}_{\alpha_{t+\Theta}} - \vec{X}_{\alpha_t}$ |
| Total walking distance | $\omega_\alpha = \sum_{t=ti_\alpha}^{to_\alpha - 1} d_{\alpha_t}$ |
| Average speed | $\Psi_\alpha = \dfrac{\omega_\alpha}{to_\alpha - ti_\alpha}$ |
| Straight-line distance | $\Omega_\alpha = \|\vec{X}_{\alpha_{to}} - \vec{X}_{\alpha_{ti}}\|$ |
| Straight-line displacement | $\vec{\Omega}_\alpha = \vec{X}_{\alpha_{to}} - \vec{X}_{\alpha_{ti}} = \sum_{t=ti_\alpha}^{to_\alpha - 1} \vec{d}_{\alpha_t}$ |
| Average walking displacement | $\vec{\xi}_\alpha = \dfrac{\sum_{t=ti_\alpha}^{to_\alpha - 1} \vec{d}_{\alpha_t}}{\rho_\alpha - 1} = \dfrac{\vec{\Omega}_\alpha}{\rho_\alpha - 1}$ |
| Variance of walking displacement | $\vec{Z}_\alpha = \dfrac{\sum_{t=ti_\alpha}^{to_\alpha - 1} (\vec{d}_{\alpha_t} - \vec{\xi}_\alpha)^2}{\rho_\alpha - 1}$ |
| Uncomfortability index | $\gamma_\alpha = \dfrac{\|\vec{Z}_\alpha\|}{\omega_\alpha}$ |
| Delay | $\lambda_\alpha = \dfrac{\omega_\alpha - \Omega_\alpha}{\omega_\alpha \cdot \Psi_\alpha}$ |

**Table 3. Traffic Flow Variables**

| Traffic Flow Variable | Formula |
|---|---|
| Flow rate | $q = \dfrac{n}{T_2 - T_1}$ |
| Time mean speed | $TMS = \dfrac{\sum_\alpha \Psi_\alpha}{n}$ |
| Space mean speed | $SMS = \dfrac{L}{\frac{1}{n}\sum_\alpha \frac{L}{\Psi_\alpha}} = \dfrac{n}{\sum_\alpha \frac{1}{\Psi_\alpha}}$ |
| Area module | $M = \dfrac{A}{n}$ |
| Moving direction | $\vec{m}_\alpha = \dfrac{\vec{X}_{\alpha_{to}} - \vec{X}_{\alpha_{ti}}}{\|\vec{X}_{\alpha_{to}} - \vec{X}_{\alpha_{ti}}\|}$ |